\documentclass[letterpaper, 10 pt, journal, twoside]{IEEEtran}

\usepackage[utf8]{inputenc}
\usepackage{hyperref}
\usepackage{url}
\usepackage{amsfonts}
\usepackage{amssymb}
\usepackage{amsmath}
\usepackage{dsfont}
\usepackage{algorithm}
\usepackage{algorithmicx}
\usepackage{xcolor}
\usepackage{comment}
\usepackage[pdftex]{graphicx}
\usepackage{physics}
\usepackage{multirow}
\usepackage{tikz}
\usepackage{ifthen}
\usepackage{adjustbox}
\usepackage{lipsum}
\usepackage{tabularx}
\usepackage{calc}
\usepackage{svg}
\usepackage{cite}
\usepackage{afterpage}
\usepackage{csquotes}
\usepackage{booktabs}
\usepackage[euler]{textgreek}
\DeclareMathOperator*{\argmin}{argmin}

\begin{document}

\title{Improving Trust Estimation in Human-Robot Collaboration Using Beta Reputation at Fine-grained Timescales}

\author{Resul Dagdanov, Milan Andrejevi\'{c}, Dikai Liu, and Chin-Teng Lin%
\thanks{This work was supported in part by the Australian Research Council (ARC) Research Hub (IH240100016).
\textit{(Corresponding author: Resul Dagdanov.)}}%
\thanks{Resul Dagdanov and Dikai Liu are with the Robotics Institute, Faculty of Engineering and Information Technology, University of Technology Sydney, Ultimo, NSW 2007, Australia
        {\tt\footnotesize Resul.Dagdanov@uts.edu.au Dikai.Liu@uts.edu.au}}%
\thanks{Milan Andrejevi\'{c} is with the Psychology Discipline, Graduate School of Health, Faculty of Health, University of Technology Sydney, Ultimo, NSW 2007, Australia
        {\tt\footnotesize Milan.Andrejevic@uts.edu.au}}%
\thanks{Chin-Teng Lin is with the Australian Artificial Intelligence Institute, School of Computer Science, Faculty of Engineering and Information Technology, University of Technology Sydney, Ultimo, NSW 2007, Australia
        {\tt\footnotesize Chin-Teng.Lin@uts.edu.au}}%
}

\maketitle

\begin{abstract}
When interacting with each other, humans adjust their behavior based on perceived trust. To achieve similar adaptability, robots must accurately estimate human trust at sufficiently granular timescales while collaborating with humans. Beta reputation is a popular way to formalize a mathematical estimation of human trust. However, it relies on binary performance, which updates trust estimations only after each task concludes. Additionally, manually crafting a reward function is the usual method of building a performance indicator, which is labor-intensive and time-consuming. These limitations prevent efficient capture of continuous trust changes at more granular timescales throughout the collaboration task. Therefore, this paper presents a new framework for the estimation of human trust using beta reputation at fine-grained timescales. To achieve granularity in beta reputation, we utilize continuous reward values to update trust estimates at each timestep of a task. We construct a continuous reward function using maximum entropy optimization to eliminate the need for the laborious specification of a performance indicator. The proposed framework improves trust estimations by increasing accuracy, eliminating the need to manually craft a reward function, and advancing toward the development of more intelligent robots.~\footnote[1] {The source code and media materials are publicly available at \url{https://github.com/resuldagdanov/robot-learning-human-trust}}
\end{abstract}

\IEEEpeerreviewmaketitle

\section{Introduction} \label{sec: introduction}
\afterpage{
    \begin{figure}[htbp]
        \centering
            \includegraphics[width=0.9271\linewidth]{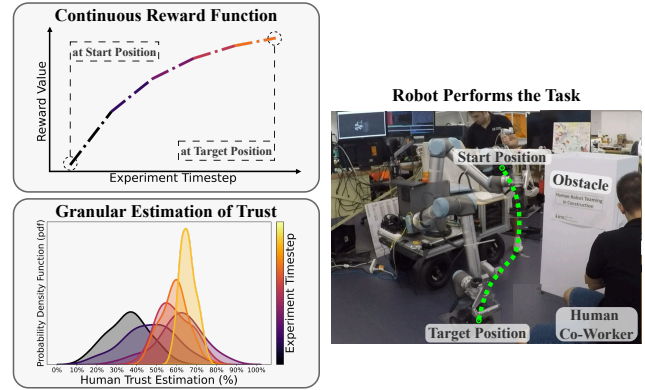}
              \caption{{\bfseries{Overview of a real-time estimation of human co-worker's trust.}} The objective of the intelligent robot is to autonomously transport tiles from a randomly chosen starting position to a target position. While performing the task, the framework continuously assigns a reward value at each timestep. The proposed framework mathematically links the reward values to the probabilistic trust estimation at fine-grained timescales. Doing so enables a real-time estimation of human trust at each timestep throughout the task.}
        \label{fig: Teaser}
    \end{figure}
}
\afterpage{
    \begin{figure*}[t!]
        \centering
            \includegraphics[width=0.9999\linewidth]{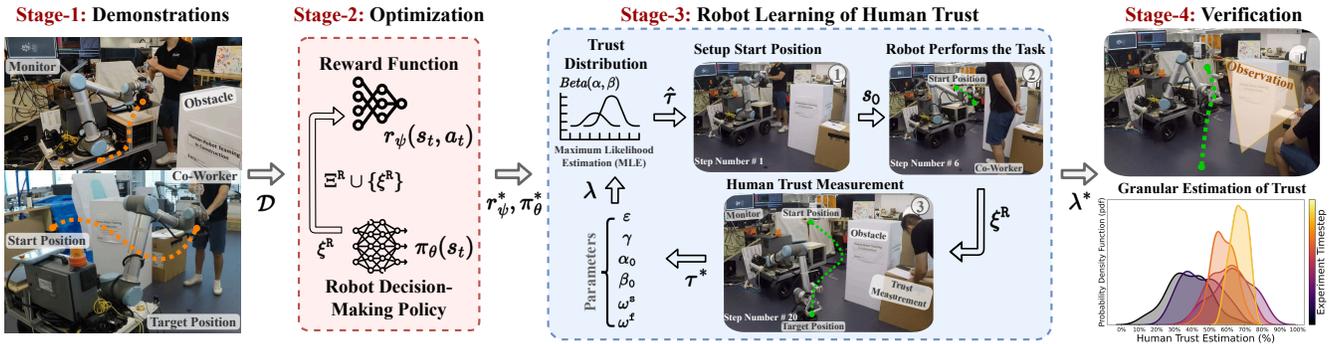}
                \caption{{\bfseries{Proposed framework for human trust estimation in HRC.}} {\bfseries{Stage-1:}} A co-worker demonstrates the task objectives to an intelligent robot by physically moving the robotic arm. {\bfseries{Stage-2:}} In this stage, a reward function, $\mathit{r_{\psi}}$, and the robot's decision-making policy, $\pi_{\theta}$, are optimized through maximum entropy optimization and behavior cloning, respectively. {\bfseries{Stage-3:}} The robot iteratively learns the trust dynamics of the human co-worker over multiple consecutive tasks. For each task, the co-worker randomly sets a starting position for the robotic arm. The robot performs the task using the optimized decision-making policy ($\pi^{}_{\theta}$). After the robot completes the task ($\mathtt{T} = 20$), the co-worker self-reports trust. The optimization objective is to minimize the difference between the measured trust ($\tau^{}$) and the estimated trust ($\hat{\tau}$) using maximum likelihood estimation. The parameters for optimization include success and failure weights, initial probability distribution parameters, the success-failure reward determination threshold, and the trust history dependency constant, collectively represented as $\lambda$ = {$\omega^\mathtt{s}$, $\omega^\mathtt{f}$, $\alpha_\mathtt{0}$, $\beta_\mathtt{0}$, $\varepsilon$, $\gamma$}. {\bfseries{Stage-4:}} In the verification stage, the robot estimates the human co-worker's trust in real-time at fine-grained timescales.}
        \label{fig: GeneralFramework}
    \end{figure*}
}
\IEEEPARstart{H}{uman} decisions are often influenced by their perceptions of how trustworthy they are perceived by others~\cite{king2005getting, ferrin2008takes}. Research in human-robot collaboration (HRC) indicates that when robots act in accordance with a human co-worker's trust, collaboration effectiveness is enhanced~\cite{lewandowsky2000dynamics, lewis2018role, gunia2024role}. However, to make trust-aware decisions, a robot needs to accurately estimate how much its co-worker trusts it~\cite{xu2015optimo, nam2020trust}.

Trust in a robot can change throughout a task, making it essential for the robot to estimate trust in real-time at fine-grained timescales. By continuously estimating trust during the task rather than only at its conclusion, the robot can adapt its behavior immediately, either enhancing or reducing trust to address the pitfalls of overtrust or undertrust~\cite{kaniarasu2013robot, de2020towards, yang2016users}.

There is growing HRC research interest in computational models to estimate human trust toward robots~\cite{wang2023human, lewis2021deep, wang2022computational, williams2023computational}. These models are based on robot performance, which is the most significant factor influencing human trust~\cite{hancock2011meta, schaefer2016measuring, yang2016users}. Furthermore, probabilistic models that capture uncertainty and bias in human subjectivity show great promise in this context~\cite{chen2018planning, nam2019models, guo2021modeling, bhat2022clustering}. The framework in this paper entails a probabilistic estimation of trust based on performance (Fig.~\ref{fig: Teaser}).

The probabilistic models proposed in~\cite{chen2018planning, nam2019models, guo2021modeling, bhat2022clustering} fail to capture the continuous changes in human trust as a robot performs a task. This limitation arises because human co-workers assess performance in a binary manner (e.g., success or failure) only after task completion, neglecting performance changes during the task. This results in a static estimation of trust dynamics, often referred to as a \enquote{snapshot} view~\cite{guo2021modeling}.

Formulating a task-specific performance function is labor-intensive, time-consuming, and requires a deep understanding of the task~\cite{biyik2023active, biyik2022learning, finn2016guided}. This process involves determining appropriate weights for the factors influencing task objectives and aligning them with desired outcomes. This limits the autonomy and adaptability of the robot to various tasks and highlights the need for a framework that enables fine-grained estimation of human trust, facilitating real-time trust-aware robot decision-making with minimal labor-intensive effort.

We propose a new framework for the accurate estimation of human trust at granular timescales, as visualized in Fig.~\ref{fig: GeneralFramework}. We construct a continuous reward function using maximum entropy optimization, which enables us to efficiently capture the underlying performance dynamics throughout the task.

This study focuses on trust estimation at fine-grained timescales. The proposed framework establishes a mathematical connection between continuous reward functions and probabilistic trust dynamics. Although the current framework does not yet incorporate robot behavior adaptation based on estimated trust, we consider fine-grained trust estimation a critical foundation for enabling real-time behavioral adaptation. To elaborate, in Section~\ref{sec: conclusion}, we outline potential strategies for trust-aware decision-making. For instance, if a human's trust declines following a task failure, the robot could communicate an apology and a commitment to improve~\cite{de2020towards, li2024trust}. Such communications may help mitigate undertrust in real-time, preventing disproportionate deterioration. Developing these behavioral adaptations requires further research, which we leave for future exploration.

\section{Literature Review} \label{sec: related_work}
\subsection{Probabilistic and Deterministic Trust Models in HRC}
A computational model is necessary for the estimation of human trust. One straightforward way is to construct trust estimation as a linear combination of performance features that influence trust in HRC~\cite{wang2022computational, wu2017toward}. Such models do not incorporate uncertainty, which is advantageous in some tasks. For example, in repetitive assembly tasks where the robot's reliability is consistent and predictable, deterministic models of trust are applied~\cite{wu2017toward, john1992trust}. However, in cases where it is important to capture human subjectivity, these models fall short. This is because human perceptions and decision-making typically involve uncertainty~\cite{fleming2024metacognition, king2005getting, ferrin2008takes}. Therefore, human perceptions of trustworthiness most likely include subjective uncertainty, which informs their decisions. Similarly, for robots to adjust their behavior based on human trust, they must capture uncertainty in their estimations.

To capture uncertainty in trust estimations, it is required to apply probabilistic models. A typical model is a dynamic Bayesian network. However, this model lacks a mathematical framework to describe how human trust stabilizes over time through repeated collaborations with the same robot~\cite{yang2017evaluating}. A more suitable alternative is the beta reputation, which has been proposed to address this limitation~\cite{guo2021modeling}, where a beta distribution offers two main advantages~\cite{josang2002beta}. Firstly, it limits an estimation interval to $0$ and $1$, creating consistency with the trust measurement scale. Secondly, this model accounts for a historical reputation by accumulating the number of successful and unsuccessful collaborations. These advantages make beta reputation suitable for the probabilistic estimation of trust.

\subsection{Robot Learning of Human Trust From Demonstrations}
To reduce the laborious workload in designing trust models, effective use of demonstrations is required. These demonstrations provide insight into how a co-worker expects a robot to perform tasks~\cite{malle2020trust}. Note that trust dynamics in HRC closely depend on the co-worker's expectations of the robot's capabilities~\cite{lee2004trust, hancock2011meta}. Thus, these demonstrations are a critical resource for modeling trust, as they reflect human expectations.

The maximum entropy optimization method can quantify the similarity mismatch between co-worker demonstrations and a robot's capabilities~\cite{ziebart2008maximum}. It was applied to construct reward and trust estimation models using demonstrations~\cite{nam2019models}. However, by clustering similar states into a fixed number of decision-making policies, this method loses flexibility in environments where decision-making parameters are not constant. Furthermore, recent work in~\cite{bhat2022clustering} has applied this method to learn personalized weights in trust estimation. Similarly, the work in~\cite{campagna2024promoting} has used the preference-based optimization~\cite{bemporad2021global} to capture individual differences. For example, one co-worker may prioritize safety, while another may emphasize a robot's speed as a critical trust indicator.

\section{Problem Formulation} \label{sec: problem_formulation}
\subsection{Markov Decision Process}
Let us consider a decision-making policy where an intelligent robot performs the task in alignment with human demonstrations. The expression for this process is a Markov decision process (MDP), denoted as $\mathcal{M} \mathrel{\mathop:}= \langle \mathcal{S}, \mathcal{A}, \mathit{r}, \mathit{f}, \mathtt{T} \rangle$. At each timestep (step number) $\mathit{t}$ in a task that concludes after a total of $\mathtt{T}$ timesteps, $\mathit{s_t}~\in~\mathcal{S}$ represents a robot state vector, and $\mathit{a_t} \in \mathcal{A}$ represents a robot action vector. An intelligent robot performs an action under the decision-making policy and then transitions to a new state $\mathit{s}_{\mathit{t} + 1}~=~\mathit{f}(\mathit{s_t},~\mathit{a_t})$ after receiving a reward $\mathit{r(\mathit{s_t},~\mathit{a_t})}$, where $\mathit{f}$ is a transition function.

\subsection{Robot Operations and Human Demonstrations}
Robot operations can take various forms, such as kinesthetic navigation, audio, and visual communication. In this work, operations performed by a robot are robotic arm manipulations under the control of a decision-making policy. As a robot performs actions, it transitions to a new state based on transition function dynamics, similar to those described in~\cite{biyik2023active, biyik2022learning}. A consecutive sequence of these transitions is a spatial trajectory, which is represented as a finite set of $\mathtt{T}$ timestep state-action pairs and notated as $\xi~=~\{(\mathit{s_1},~\mathit{a_1}),~(\mathit{s_2},~\mathit{a_2}),~\ldots,~(\mathit{s_\mathtt{T}},~\mathit{a_\mathtt{T}})\}~\in~\Xi$. To simplify the notation, as adopted from~\cite{biyik2023active}, a trajectory can be denoted in a compact form as $\varsigma~=~(\mathit{s_1},~\mathit{a_1},~\mathit{a_2},~\ldots,~\mathit{a_\mathtt{T}})$.

Human demonstrations refer to samples where a co-worker physically shows the robot how to perform a task by manipulating the robotic arm, like the kinesthetic interactions described in~\cite{wang2022computational}. The notation for a dataset of $N$ demonstrations is $\mathcal{D}~=~\{\xi^{\mathtt{H}}_{1},~\xi^{\mathtt{H}}_{2},~\ldots,~\xi^{\mathtt{H}}_{N}\}$, where $\xi^{\mathtt{H}}$ is a sample trajectory demonstration by a human co-worker.

\subsection{Human Trust Definition} \label{sec: HumanTrustFormulation}
In this paper, a human is the trustor, and a robot is the trustee. The concept of human trust is defined in~\cite{lee2004trust} as~\enquote{the attitude that an agent will help achieve an individual’s goals in a situation characterized by uncertainty and vulnerability}. This definition of trust aligns with the engineering aspects and the goal-oriented nature of a robot in HRC.

In HRC literature~\cite{nam2020trust}, models of human trust estimations are categorized into relation-based and performance-based models. Performance-based models estimate trust primarily based on the capability and reliability of a robot. In contrast, relation-based models use data on the societal and ethical norms of a human as a trust estimation feature. This paper presents a performance-based model of human trust because the engineering objective is to enhance the ability of a robot to perform physical tasks with high performance.

The performance of the robot in HRC is the most dominant factor affecting human trust in the robot~\cite{hancock2011meta, schaefer2016measuring, yang2016users}. Trust depends on the successful and unsuccessful reputation of collaboration with the robot. In this paper, a success metric of collaboration is the compatibility between human co-worker expectations and robot capabilities, which serves as a key indicator of changes in trust~\cite{malle2020trust}. Based on these findings, we present a performance-based model of trust.
\afterpage{
    \begin{figure}[htbp]
        \centering
            \includegraphics[width=0.9874\linewidth]{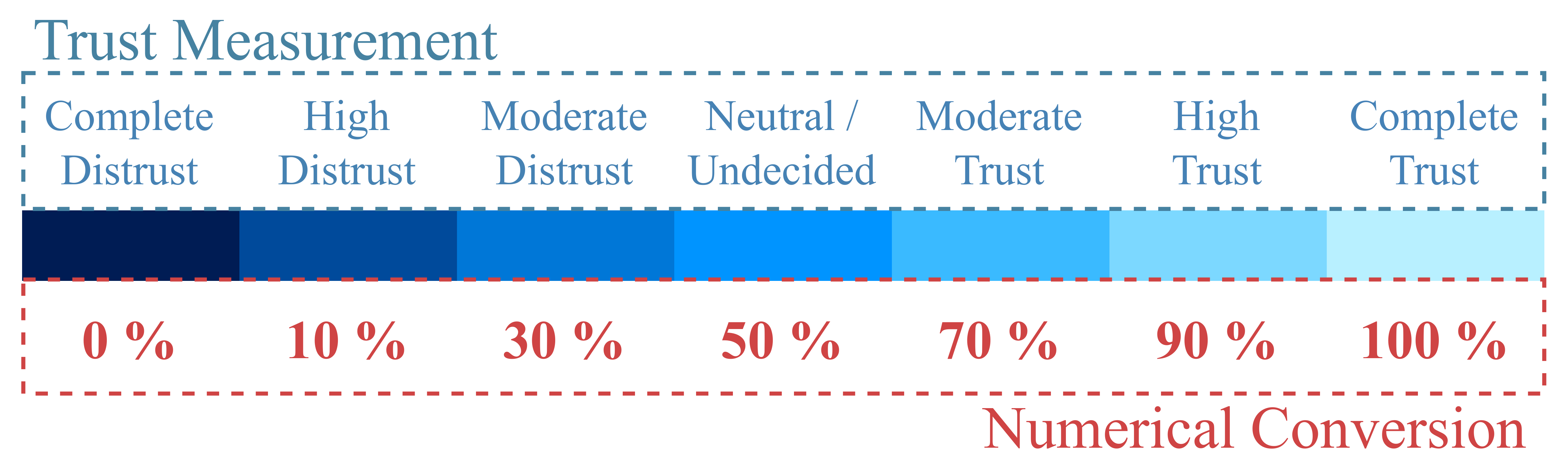}
                \caption{{\bfseries{Measurement of human trust and numerical conversion.}} We utilize a 7-point Likert scale to obtain human co-worker trust measurements, as this self-reporting technique is well-established in the literature~\cite{nam2020trust, schaefer2016measuring}.}
        \label{fig: TrustScale}
    \end{figure}
}

The assumption in this paper is that the reward function represents the robot's task performance. To examine the relationship between the reward function and task performance, Section~\ref{sec: experimental_evaluation} provides an experimental analysis.

At the end of each experiment, a human co-worker self-reports their trust using a 7-point Likert scale. As shown in Fig.~\ref{fig: TrustScale}, we numerically interpret the trust values as percentages. This conversion to the percentage scale is kept constant and is symmetrical around the midpoint (i.e., $50$\%, which corresponds to the \enquote{Neutral} level of trust on the scale). This allows the robot to estimate trust based on numerical data.

\subsection{Robot Decision-Making Policy} \label{sec: BehaviorCloning}
To enhance the robot's generalization capabilities, particularly in high-dimensional action spaces, a learning-based method for robot decision-making needs to be implemented. One common approach to doing this is to maximize the similarity between expert demonstrations and robot operations, also known as behavior cloning (BC). This approach comes with a major limitation. Mimicking human actions may result in a suboptimal decision-making policy for the robot, as humans do not always act optimally. However, there is also a big advantage to this approach. As discussed in Section~\ref{sec: HumanTrustFormulation}, the similarity mismatch between co-worker demonstrations and robot operations can serve as a key indicator of changes in trust, which can be captured via the reward function. For this reason, in our framework, the human co-worker's demonstrations are treated as optimal (i.e., expert).

A common practice in formulating a robot decision-making policy is to assume that human demonstrations resemble a Gaussian distribution~\cite{finn2016guided, biyik2023active}. Consequently, a robot policy is $\pi_{\theta}(\mathit{a_t}~\hspace{-2.0mm}~\mid~\hspace{-2.0mm}~\mathit{s_t})~\sim~\mathcal{N}(\mu_{\theta}(\mathit{s_t}), \sigma_{\theta}(\mathit{s_t})^2)$, where $\mu_{\theta}(\mathit{s_t})$ and $\sigma_{\theta}(\mathit{s_t})$ denote the mean and variance of the policy distribution, respectively. In this manner, $\mu_{\theta}$ and $\sigma_{\theta}$ serve as parameters of a nonlinear neural network.
\begin{equation}
        \pi_{\theta}^* = \argmin_{\theta} \frac{1}{2} \cdot \hspace{-3.0mm} \underset{\substack{\xi^{\mathtt{H}} \sim \mathcal{D} \\ (\mathit{s_t}, \mathit{a_t}) \sim \xi^{\mathtt{H}}}}{\mathbb{E}} \hspace{-1.5mm} \left[ \eta \cdot \text{log} \hspace{0.5mm} \sigma_{\theta}(\mathit{s_t})^2 \scalebox{0.99}{\(+\)} \frac{(\mathit{a_t} \scalebox{0.99}{\(-\)} \mu_{\theta}(\mathit{s_t}))^2}{\sigma_{\theta}(\mathit{s_t})^2} \right]
    \label{eqn: bc_loss}
\end{equation}

Eq.~\ref{eqn: bc_loss} represents a minimization objective function for optimizing a robot decision-making policy. The $\eta$ coefficient acts as a scaling factor for regularizing a policy variance. When $\eta$ is large, a policy optimization prioritizes minimizing a decision-making uncertainty. In contrast, when $\eta$ is small, a robot performs more stochastic actions, leading to increased exploration. Section~\ref{sec: OptimizingRobotPolicy} provides a detailed description of how $\eta$ varies throughout this data-driven optimization process, which occurs in Stage-2 of Fig.~\ref{fig: GeneralFramework}.

\subsection{Reward Function Optimization} \label{sec: InverseReinforcementLearning}
Formulating a reward function that accurately captures the context-dependent performance of a task is laborious and time-consuming~\cite{biyik2023active, biyik2022learning, finn2016guided}. This complexity arises from the challenge of determining appropriate weights that align the reward function with desired outcomes. One solution to this problem is to use a data-driven approach, specifically maximum entropy (MaxEnt) optimization~\cite{ziebart2008maximum}, to construct a reward function that accurately captures the performance of robotics applications~\cite{finn2016guided, swamy2023inverse}. So, to eliminate the need for laborious performance specifications, we use MaxEnt optimization to construct a continuous reward function.

In the early learning epochs of BC in Stage-2 of Fig.~\ref{fig: GeneralFramework}, a robot performs suboptimal actions. As the optimization process in Eq.~(\ref{eqn: bc_loss}) proceeds, a decision-making policy gets closer to an optimal policy. The main idea behind MaxEnt optimization is to iteratively sample trajectories from $p(\xi)~\sim~\text{exp}(\mathcal{R(\xi)})$. The goal is to match features between robot trajectories~$\Xi^{\mathtt{R}}$ and human demonstrations~$\mathcal{D}$.

The proposed framework does not focus on learning a robot policy by maximizing cumulative rewards. Instead, the aim of this paper is to learn a robot policy through BC by enabling it to mimic demonstrations (see Section~\ref{sec: BehaviorCloning}).
\begin{equation}
        \mathcal{R_{\psi}(\xi)} = \frac{1}{\mathtt{T}} \hspace{-1mm} \sum_{(\mathit{s_t}, \mathit{a_t}) \in \xi} \hspace{-1.5mm} \mathit{r_{\psi}(\mathit{s_t}, \mathit{a_t})} \text{  } \mid \text{  } \mathit{r_{\psi}} \text{} : \text{ } (\mathcal{S}, \mathcal{A}) \rightarrow \mathbb{R}^{\left[ \text{-1, 1} \right]}
    \label{eqn: reward_function}
\end{equation}

Formulating a linear reward function in high-dimensional environments is challenging and often impractical. A nonlinear approach is necessary to construct a reward function for these environments. A widely adopted method for capturing performance is the use of nonlinear neural networks, which provide flexibility and adaptability in data-driven solutions~\cite{finn2016guided}. Therefore, we construct a reward function in Eq.~(\ref{eqn: reward_function}) using a neural network with parameters~$\psi$.
\begin{equation}
        \mathcal{L_{\text{MaxEnt}}} = \scalebox{0.8}{\(-\)} \underset{\scriptscriptstyle \xi^{\mathtt{H}} \sim \mathcal{D}}{\mathbb{E}} \Big[ \text{log} \hspace{0.5mm} p(\xi^{\mathtt{H}} | \psi) \Big] = \scalebox{0.8}{\(-\)} \underset{\scriptscriptstyle \xi^{\mathtt{H}} \sim \mathcal{D}}{\mathbb{E}} \left[ \text{log} \frac{\text{exp}(\mathcal{R_{\psi}})}{\mathcal{Z}(\psi)} \right]
    \label{eqn: max_ent_loss}
\end{equation}

Eq.~(\ref{eqn: max_ent_loss}) is a loss function of MaxEnt optimization. There are infinitely many discrete possible states and actions for the background partition function $\mathcal{Z}(\psi)~=~\int~\text{exp}(\mathcal{R_{\psi}})~d\xi$ to calculate when $\mathcal{S}$ and $\mathcal{A}$ are both continuous.
\begin{equation}
        \mathcal{Z}(\psi; \theta) \approx \frac{1}{M} \sum_{\xi^{\mathtt{R}}_j \in \Xi^{\mathtt{R}}} \left[ \frac{\text{exp}(\mathcal{R_{\psi}}(\xi^{\mathtt{R}}_j))}{p(\xi^{\mathtt{R}}_j ; \theta)} \right]
    \label{eqn: partition_function}
\end{equation}

A stochastic sampling-based method is a popular technique for approximating~$\mathcal{Z}(\psi; \theta)$, as proposed in~\cite{finn2016guided}. This method, as shown in Eq.~(\ref{eqn: partition_function}), approximates close to the expectation of the negative log-likelihood loss in Eq.~(\ref{eqn: max_ent_loss}). In this paper, $\mathcal{D}$ and $\Xi^{\mathtt{R}}$ are the sets of $N$ demonstrations and $M$ robot trajectories, respectively. Note that it is common practice to generate robot trajectories from $p_{\theta}(\xi)$~\cite{swamy2023inverse, biyik2023active}.

To address the exploration-exploitation dilemma inherent in a decision-making policy and to enhance the generality of a reward function by approximating $\mathcal{Z}(\psi; \theta)$ in Eq.~(\ref{eqn: partition_function}), this paper demonstrates an application of a dynamically annealing/interpolating linear weight $\eta$ (see Eq.~(\ref{eqn: bc_loss})). Section~\ref{sec: OptimizingRobotPolicy} comprehensively explains the rationale behind this choice.

\section{Probabilistic Framework for Granular Estimation of Human Trust} \label{sec: learning_trust}
This section outlines the mathematical foundations for robot learning of human trust in a probabilistic manner. The divergence between human expectations and robot actions in HRC are key factors influencing the dynamics of human trust~\cite{nam2020trust, hancock2011meta, malle2020trust}. Accordingly, this section provides a mathematical framework for the estimation of human trust at granular timescales based on a continuous reward function.

\subsection{Robot Decision-Making Policy and Reward Function} \label{sec: OptimizingRobotPolicy}
The denominator in Eq.~(\ref{eqn: partition_function}) represents the MDP collection of $\pi_{\theta}(\mathit{a_t} \mid \mathit{s_t}, \mathit{s}_{t-1}, \dots)$, which denotes the probability of taking action $\mathit{a_t}$ at state $\mathit{s_t}$ according to the decision-making policy $\pi_{\theta}$. Finding the exact value of the background partition function $\mathcal{Z}(\psi; \theta)$ with $p_{\theta}(\xi)$ is infeasible, especially when the ideal reward parameters~$\psi^{*}$ are unknown.
\begin{equation}
        \eta = \eta_{min} + \frac{k}{K} \cdot \left( \eta_{max} - \eta_{min} \right)
    \label{eqn: interpolation}
\end{equation}

In~\cite{finn2016guided}, researchers employed an iterative method for a decision-making policy that explored the task environment by performing random actions. To minimize the frequency of the random actions,~\cite{swamy2023inverse} investigated the use of annealing and interpolation methods to determine the exploration weight factor. The framework in this paper utilizes a linear interpolation method to find dynamic importance weight~$\eta$. The method is given in Eq.~(\ref{eqn: interpolation}), where $k$ is the current epoch and $K$ is the maximum number of learning epochs.

In order for a data-driven reward function to represent task performance, it is necessary for a robot to explore the task environment widely via random actions. During the early epochs of learning, a decision-making policy under the objective function in Eq.~(\ref{eqn: interpolation}) prioritizes a wider exploration of the task environment. Consequently, $\eta$ gradually increases as learning epochs increase. The selection of the hyperparameters $\eta_{\text{min}} = 0.05$ and $\eta_{\text{max}} = 1.00$ in Eq.~(\ref{eqn: interpolation}) is based on problem-specific trials. It is important to note that increasing $\eta_{\text{max}}$ could lead to a less generalized reward function.

The objective is to gradually increase the loss function variance (uncertainty) factor during the learning process of a robot decision-making policy. This strategy ensures the robot exhibits significant uncertainty but explores more in the early policy and reward learning epochs. Adequate exploration is crucial for achieving a generalizable reward function in MaxEnt optimization. This process of learning a decision-making policy takes place in Stage-2 of Fig.~\ref{fig: GeneralFramework}.
\afterpage{
    \begin{figure}[htbp]
        \centering
            \includegraphics[width=0.9242\linewidth]{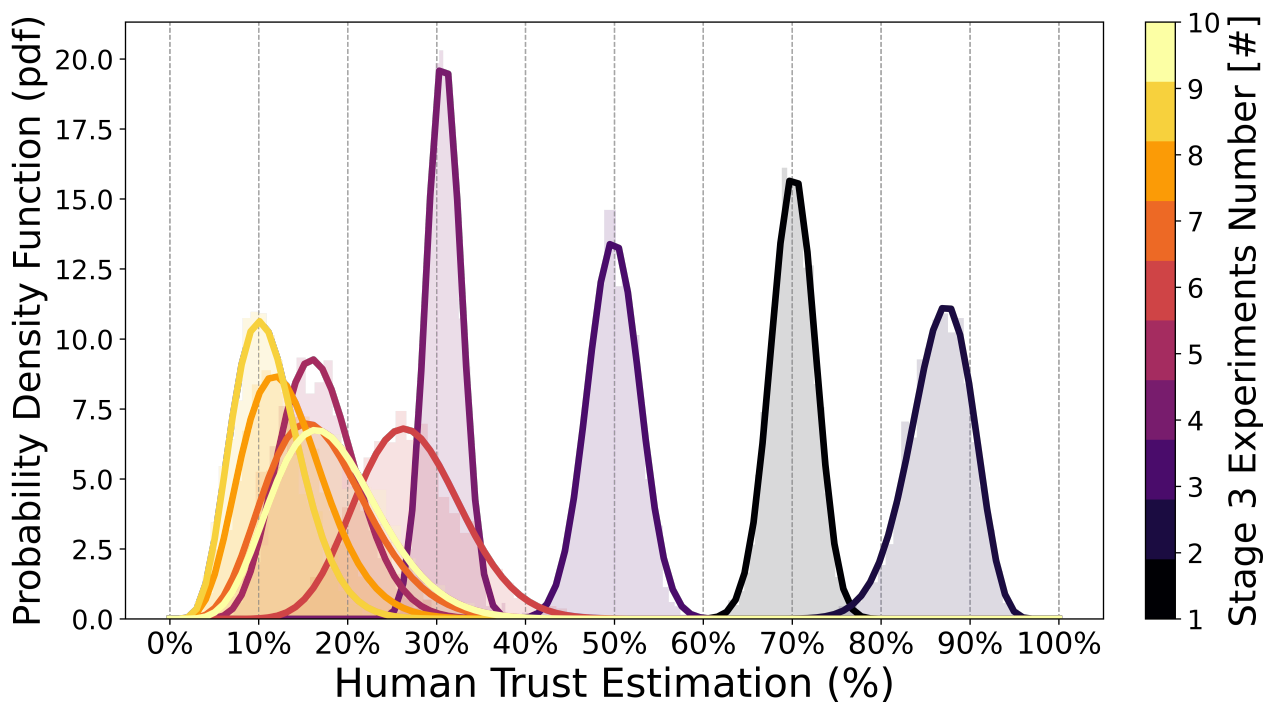}
                \caption{{\bfseries{Trust estimation dynamics during Stage-3 experiments.}} Trust estimations show narrower distributions in earlier experiments (darker shades), reflecting lower variance due to limited measurements. As experiments progress (lighter shades), distributions shift toward lower trust regions, and variance increases. This characteristic results from lower self-reported trust in Stage-3, which expanded the range of trust measurements (see Fig.~\ref{fig: LearningExperimentReward}).}
        \label{fig: LearningExperimentDistributions}
    \end{figure}
}

The collection of human demonstration data occurs in Stage-1 of Fig.~\ref{fig: GeneralFramework}. Subsequently, the learning processes for the reward function and robot decision-making policy occur in Stage-2. The total number of human demonstrations remains constant throughout the proposed framework. The parameters $\theta^{*}$ and $\psi^{*}$ represent the optimized decision-making policy and reward function obtained through MaxEnt optimization and BC, respectively. Once optimized, $\theta^{*}$ and $\psi^{*}$ remain fixed. Therefore, for a given state $\mathit{s_t}$, the framework ensures that the reward value $\mathit{r_{\psi^{*}}}$ is reproducible and the decision-making policy $\pi_{\theta^{*}}$ always takes action $\mathit{a_t}$.

\subsection{Beta Reputation Model at Fine-Grained Timescales} \label{sec: ModelingHumanTrustBehavior}
An important theoretical part of the proposed framework is an estimation of human trust by updating a beta probability distribution at each timestep of the task.

Stage-3 of Fig.~\ref{fig: GeneralFramework} involves a human co-worker testing the capability of a decision-making policy and a reward function. Each complete cycle in Stage-3 represents one experiment, after which self-reported trust is collected based on a 7-point Likert scale shown in Fig.~\ref{fig: TrustScale}. While a human co-worker reports trust at the end of each task, a reward function continuously assigns reward values to each state-action pair. This aspect enables granularity in human trust estimation.

\subsubsection{Probability Distribution}
A beta reputation system~\cite{josang2002beta} is a beta probability distribution. It is a common choice to estimate trust probabilistically~\cite{guo2021modeling, bhat2022clustering, chen2018planning}, as it captures subjective uncertainty and variability of human trust.
\begin{equation}
        \tau_{\mathit{q}}(\mathit{s_t}, \mathit{a_t}) \sim Beta(\alpha_{\mathit{n}}, \beta_{\mathit{m}})
    \label{eqn: trust_probability}
\end{equation}

Eq.~(\ref{eqn: trust_probability}) is a formulation of a continuous probability distribution to represent human co-worker's trust $\tau_{\mathit{q}}(\mathit{s_t}, \mathit{a_t})$ at robot state $\mathit{s_t}$ and robot action $\mathit{a_t}$ at timestep $\mathit{t}~\leq~\mathtt{T}$ of trajectory $\xi^{\mathtt{R}}$. $\mathit{q}$ is the total number of timesteps in all tasks.
\begin{equation}
        \hat{\tau}_{\mathit{q}}(\mathit{s_t}, \mathit{a_t}) \doteq \underset{(\mathit{s_t}, \mathit{a_t}) \sim \xi^{\mathtt{R}}}{\mathbb{E}} \Big[ \tau_{\mathit{q}}(\mathit{s_t}, \mathit{a_t}) \Big] = \frac{\alpha_{\mathit{n}}}{\alpha_{\mathit{n}} + \beta_{\mathit{m}}}
    \label{eqn: trust_value}
\end{equation}
\afterpage{
    \begin{figure}[htbp]
        \centering
            \includegraphics[width=0.9127\linewidth]{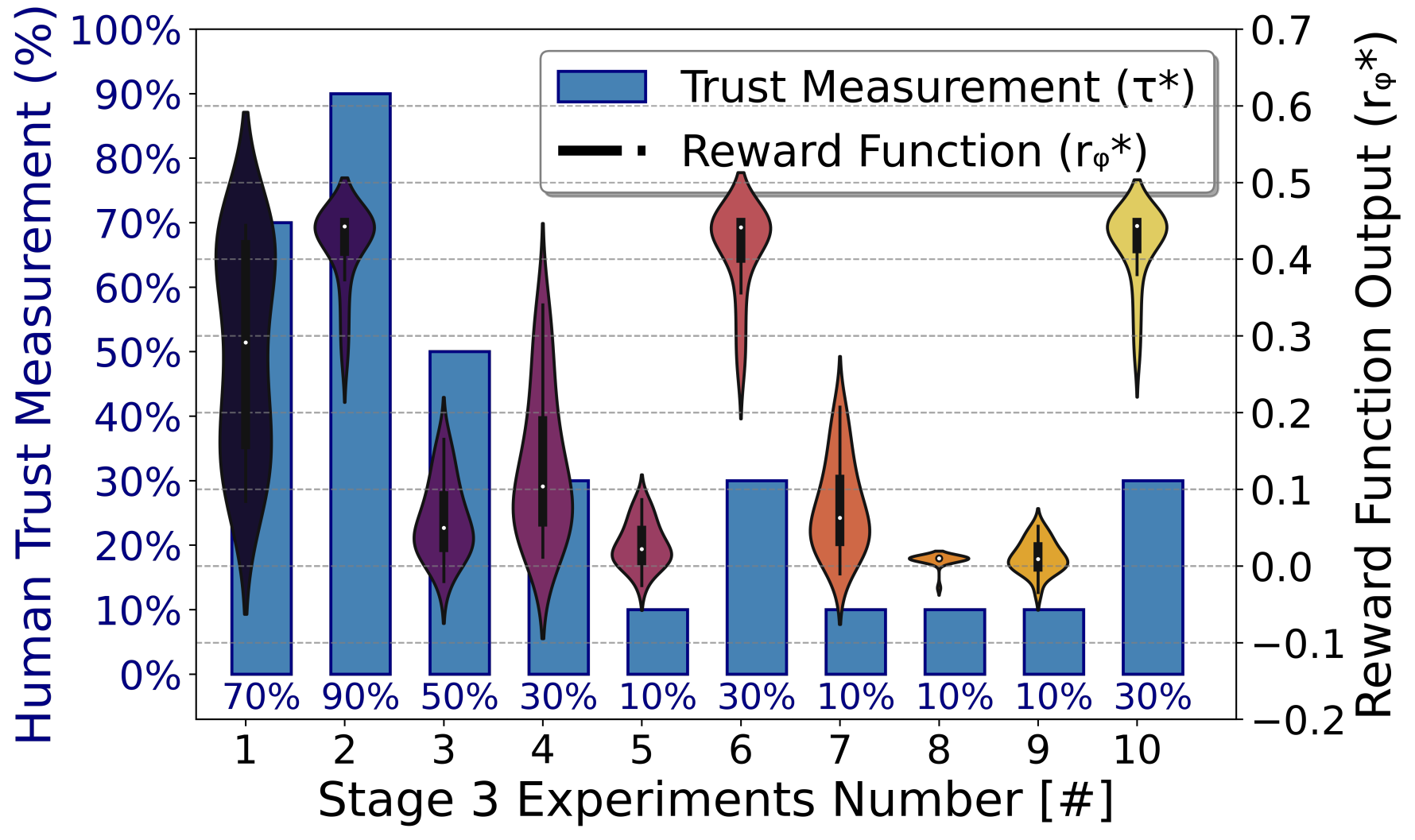}
                \caption{{\bfseries{Relationship between self-reported trust and reward in Stage-3.}} Violin plots illustrate the distribution of reward values. Trust is self-reported by the human using a constant scale (Fig.~\ref{fig: TrustScale}). Although experiments 6 and 10 had high rewards, trust remained lower than in experiment 2. This suggests that prior interactions influenced current trust, which underscores the importance of incorporating historical context into trust estimation.}
        \label{fig: LearningExperimentReward}
    \end{figure}
}

\subsubsection{Beta Distribution Parameters}
Through the experience of a human co-worker with a robot, this paper presents the mathematics of updating $\alpha_{\mathit{n}}$ and $\beta_{\mathit{m}}$ of a beta probability distribution based on a reward function $\mathit{r_{\psi^*}} \mathrel{\mathop:}= \mathit{r_{\psi^*}}(\mathit{s_t}, \mathit{a_t}) \mid (\mathit{s_t}, \mathit{a_t}) \in \xi^{\mathtt{R}}$ output at each timestep $\mathit{q}$. In Eq.~(\ref{eqn: beta_parameters}), the subscripts $\mathit{n}$ and $\mathit{m}$ indicate the total number of successful and unsuccessful state-action counts, respectively.
\begin{equation}
    \begin{aligned}
        \alpha_{\mathit{n}} &=
            \begin{cases}
                \sum\limits_{i=0}^{n \scalebox{0.75}{\(-\)} 1} \left( \gamma^{\mathit{i}} \cdot \alpha_{\mathit{n} \scalebox{0.75}{\(-\)} \mathit{i} \scalebox{0.75}{\(-\)} 1} \right), & \text{if } \mathit{r_{\psi^*}} \leq \varepsilon \\
                \sum\limits_{i=0}^{n \scalebox{0.75}{\(-\)} 1} \left( \gamma^{\mathit{i}} \cdot \alpha_{\mathit{n} \scalebox{0.75}{\(-\)} \mathit{i} \scalebox{0.75}{\(-\)} 1} \right) + \omega_{\mathit{n}}^\mathtt{s} \cdot \mathit{r_{\psi^*}}, & \text{if } \mathit{r_{\psi^*}} > \varepsilon
            \end{cases} \\
        \beta_{\mathit{m}} &= 
            \begin{cases}
                \sum\limits_{j=0}^{m \scalebox{0.75}{\(-\)} 1} \left( \gamma^{\mathit{j}} \cdot \beta_{\mathit{m} \scalebox{0.75}{\(-\)} \mathit{j} \scalebox{0.75}{\(-\)} 1} \right), & \hspace{-2.75mm} \text{if } \mathit{r_{\psi^*}} > \varepsilon \\
                \sum\limits_{j=0}^{m \scalebox{0.75}{\(-\)} 1} \left( \gamma^{\mathit{j}} \cdot \beta_{\mathit{m} \scalebox{0.75}{\(-\)} \mathit{j} \scalebox{0.75}{\(-\)} 1} \right) + \omega_{\mathit{m}}^\mathtt{f} \cdot \mathrm{e}^{\lvert \mathit{r_{\psi^*}} \rvert}, & \hspace{-2.75mm} \text{if } \mathit{r_{\psi^*}} \leq \varepsilon
            \end{cases}
    \end{aligned}
    \label{eqn: beta_parameters}
\end{equation}
\afterpage{
    \begin{figure*}[t!]
        \centering
            \includegraphics[width=0.9940\linewidth]{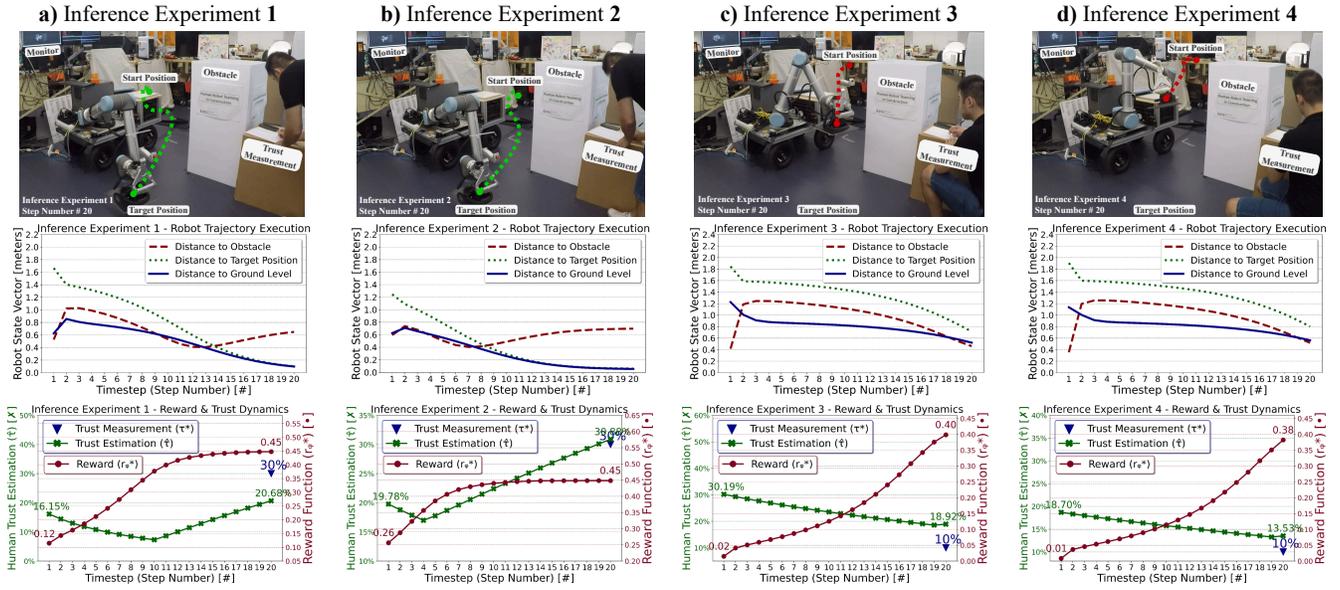}
                \caption{{\bfseries{Evaluation of the proposed framework.}} Each column (a-d) represents distinct and consecutive verification experiments in Stage-4 of Fig.~\ref{fig: GeneralFramework}. The results show four tests to evaluate human trust estimation accuracy and observe corresponding reward function values. These experiments highlight the granularity and history dependency of human trust estimation. After each experiment, the trust estimation builds on the previous experiment, reflecting the cumulative nature of human trust. In contrast, reward function values are independent of collaboration history. Experiments a-b were successful; the robot end-effector reached the target position. Experiments c-d were unsuccessful; the robot end-effector failed to reach the target position. Note that human trust measurements are discrete, but human trust estimation is at granular timescales, which dynamically change at each timestep of the experiments.}
        \label{fig: InferenceResults}
    \end{figure*}
}

Unlike~\cite{wang2022computational}, which models trust as a deterministic linear function of performance, the proposed framework uses a probabilistic approach. This allows for trust estimation at granular timescales with a more dynamic representation.

The works in~\cite{nam2020trust, frey2024interpersonal} highlight that human trust is history-dependent. To account for this characteristic of human trust, updating the $\alpha_{\mathit{n}}$ and $\beta_{\mathit{m}}$ parameters with a weighted aging technique is chosen, as shown in Eq.~(\ref{eqn: beta_parameters}). This technique includes a discount factor $(0 < \gamma \leq 1)$, reflecting the importance of previous timestep distribution parameters ($\alpha_{\mathit{n} - 1}$, $\beta_{\mathit{m} - 1}$). The primary reason for this interpretation is that human trust is cumulative and interaction history-dependent~\cite{john1992trust}. Furthermore, the assumption in this paper is that~$\mathit{r_{\psi^*}}~>~\varepsilon$ represents success, whereas $\mathit{r_{\psi^*}}~\leq~\varepsilon$ denotes failure. Therefore, in Eq.~(\ref{eqn: beta_parameters}), the constant number notation of the success factor is~$\omega_{\mathit{n}}^\mathtt{s}$ and the failure reward factor is~$\omega_{\mathit{m}}^\mathtt{f}$. Overall, trust estimation depends on the parameter set $( \alpha_\mathtt{0}, \beta_\mathtt{0}, \omega_{\mathit{n}}^\mathtt{s}, \omega_{\mathit{m}}^\mathtt{f}, \varepsilon, \gamma )$ and reward function $\mathit{r_{\psi^*}}(\mathit{s_t}, \mathit{a_t})$.

The mean of the beta distribution in Eq.~(\ref{eqn: trust_value}) estimates human trust at a given state $\mathit{s_t}$ and action $\mathit{a_t}$. Unlike~\cite{guo2021modeling}, the proposed framework updates the trust distribution at each timestep rather than only at the end of the task. Therefore, this results in an estimation at more fine-grained timescales.

\subsubsection{Maximum Likelihood Estimation} \label{sec: MaximumLikelihoodEstimation}
In Stage-3 of Fig.~\ref{fig: GeneralFramework}, the iterative process improves the mapping between reward values and human trust measurements $\tau_{\mathit{q}}^*$ using a maximum likelihood estimation (MLE) with a parameter set of $\lambda = \{ \alpha_\mathtt{0}, \beta_\mathtt{0}, \omega_{\mathit{n}}^\mathtt{s}, \omega_{\mathit{m}}^\mathtt{f}, \varepsilon, \gamma \}$. Eq.~(\ref{eqn: beta_parameters}) shows a threshold parameter $\varepsilon$ in the non-differentiable piece-wise function for the updates at granular timescales. We use a popular derivative-free differential evolution method~\cite{storn1997differential} to optimize $\lambda$. The objective is to minimize the negative log-likelihood between $\tau_{\mathit{q}}^*$ and $\hat{\tau}_{\mathit{q}}$ after each experiment in Stage-3.

\subsection{Major Empirical Findings on Trust Dynamics} \label{sec: ExperientialObservations}
\subsubsection{History Dependency} Trust at the previous timestep, $\tau_{\mathit{q} - 1}$, influences trust at the immediate next timestep, $\tau_{\mathit{q}}$. Research in~\cite{john1992trust} highlights this history-dependent nature of trust. The proposed beta reputation in Eq.~(\ref{eqn: beta_parameters}) mathematically captures this characteristic of trust with an aging factor~$\gamma$.

\subsubsection{Impact of Adverse Experiences} In inference experiment 3, the robot failed to bring the tile to the target position, as shown in Fig.~\ref{fig: InferenceResults}. Trust estimation at the first timestep of this experiment was $30.19$\%, and it continued to decrease, reaching $18.92$\% by the end of the experiment. For verification, human trust was self-reported as \enquote{Moderate Distrust} after inference experiment 2 and \enquote{High Distrust} at the end of inference experiment 3. This suggests that the human trust was affected by an adverse experience. This observation aligns with previous findings in the literature, which indicate that a negative experience significantly impacts trust~\cite{yang2016users}.

\subsubsection{Convergence of Human Trust} When $\mathit{n}$~,~$\mathit{m}$~$\rightarrow$~$\infty$, $\sum_{i=0}^{n \scalebox{0.75}{\(-\)} 1} \left( \gamma^{\mathit{i}} \cdot \alpha_{\mathit{n} \scalebox{0.75}{\(-\)} \mathit{i} \scalebox{0.75}{\(-\)} 1} \right)$ and $\sum_{j=0}^{m \scalebox{0.75}{\(-\)} 1} \left( \gamma^{\mathit{j}} \cdot \beta_{\mathit{m} \scalebox{0.75}{\(-\)} \mathit{j} \scalebox{0.75}{\(-\)} 1} \right)$ in Eq.~(\ref{eqn: beta_parameters}) diminish due to a discount factor $(0 < \gamma \leq 1)$. As a result, the contribution of earlier interactions gradually becomes negligible. Additionally, the terms $\omega_{\mathit{n}}^\mathtt{s} \cdot \mathit{r_{\psi^*}}$ and $\omega_{\mathit{m}}^\mathtt{f} \cdot \mathrm{e}^{\lvert \mathit{r_{\psi^*}} \rvert}$ in Eq.~(\ref{eqn: beta_parameters}) only shift the beta distribution by a constant factor without introducing instability in the proposed beta reputation. Consequently, because parameters $\theta^*$ and $\psi^*$ are constant in Stage-3 and Stage-4, trust estimations converge to a stable condition after repeated collaboration with the same robot, as stated in the previous research in~\cite{yang2017evaluating}.

\section{Experimental Evaluation and Results} \label{sec: experimental_evaluation}
We conducted a case study to evaluate our framework for fine-grained trust estimation in a tiling task, where a robot transported a tile while avoiding an obstacle (Fig.~\ref{fig: Teaser}). The experiment included two stages: Stage-3 with iterative model updates and Stage-4 using a fixed model. One participant (a 25-year-old male with prior technical experience in HRC) completed multiple experiments, and after each experiment, this participant self-reported trust using the scale in Fig.~\ref{fig: TrustScale}.

The state vector ($\mathit{s_t} \in \mathcal{S}$) included distances to the obstacle, the ground, and the target position, similar to~\cite{biyik2023active, biyik2022learning}. The action vector ($\mathit{a_t} \in \mathcal{A}$) represented the x, y, and z positions of the robot’s end-effector at each timestep~$\mathit{t}$.

\subsection{Reward Function and Human Trust Measurements} \label{sec: RewardTrustRelationship}
In each Stage-3 cycle (Fig.~\ref{fig: GeneralFramework}), a human randomly set the robot’s initial position at the start location. The robot then executed a policy $\pi_{\theta}(\mathit{a_t}~\mid~\mathit{s_t})$ to transport a tile to a target location while avoiding an obstacle.

Trust measurements did not always align with reward values (see Fig.~\ref{fig: LearningExperimentReward}), as the reward function does not account for the history of HRC. In contrast, the beta reputation captures history-dependent trust through cumulative interactions.

With the proposed framework, the robot estimated decreasing levels of trust in the subsequent experiments of Stage-3 (see Fig.~\ref{fig: LearningExperimentDistributions}). This downward trend corresponds to the low reward values and associated trust measurements in Fig.~\ref{fig: LearningExperimentReward}. The distribution widens with increasing experiment number in Fig.~\ref{fig: LearningExperimentDistributions}, while the opposite trend appears in Fig.~\ref{fig: InferenceExperimentsComparison}. This discrepancy arises because, during Stage-3, trust estimation parameters are updated after each experiment, whereas in Stage-4, the parameters are constant.
\afterpage{
    \begin{figure}[htbp]
        \centering
            \includegraphics[width=0.8911\linewidth]{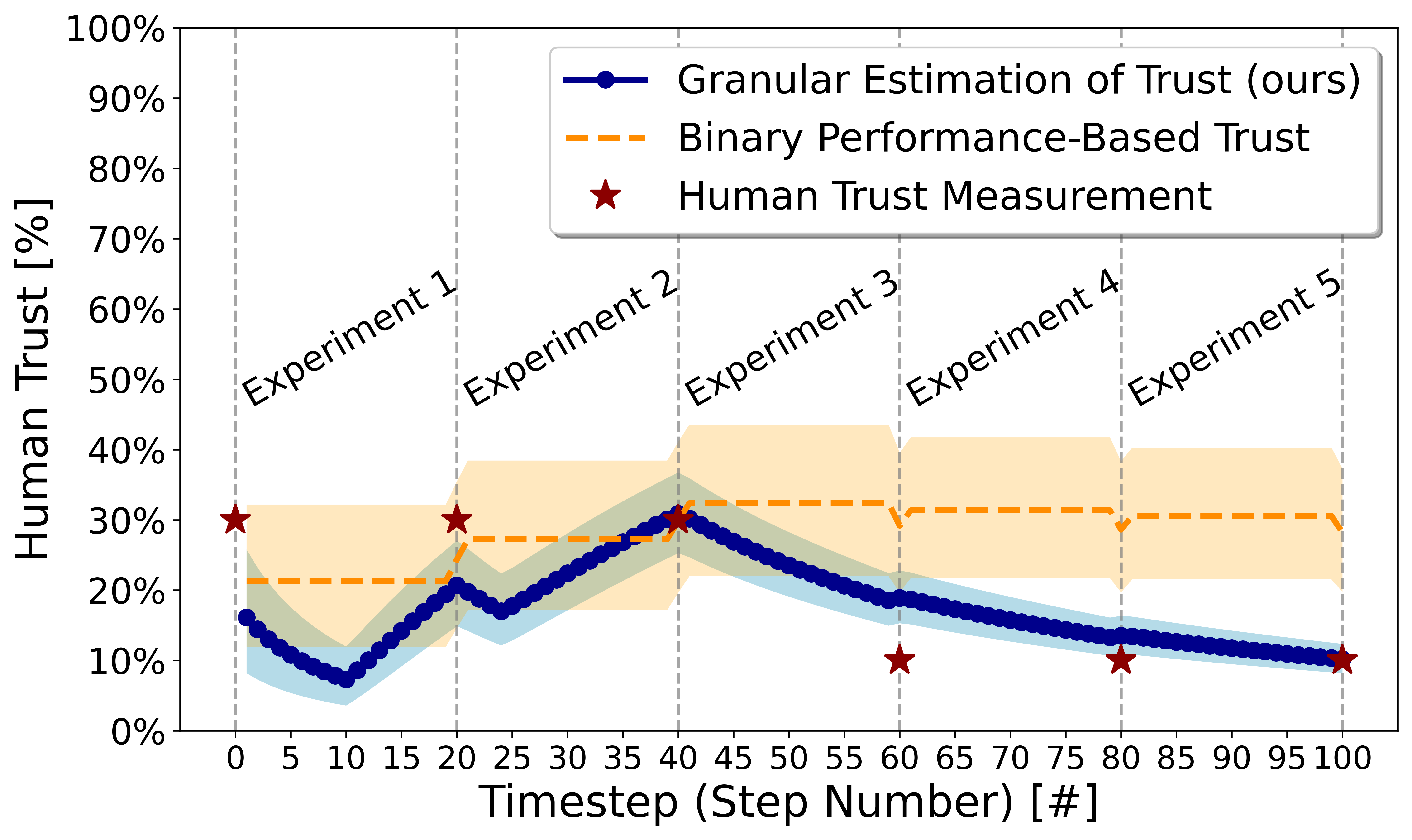}
                \caption{{\bfseries{Comparison of trust estimation models.}} The binary performance-based model only updates at the end of each experiment, as it relies on the overall success or failure of the task, resulting in a static trust estimation that remains unchanged throughout the experiment.}
        \label{fig: InferenceExperimentsComparison}
    \end{figure}
}

\subsection{Analysis of Verification Results} \label{sec: ResultsAnalysis}
In the Stage-4 experiments, the robot successfully completed the task in experiments 1 and 2 but failed in experiments 3 and 4. As shown in Fig.~\ref{fig: InferenceResults}, its failure to bring a tile to the target location led to lower trust measurements. Trust estimation also declined in experiments 3 and 4, which aligned with reduced reward values. These findings align well with the assumption presented in Section~\ref{sec: HumanTrustFormulation} that human trust is primarily affected by the robot's performance.

\subsubsection{History Dependency}
At the end of inference experiment 1, trust was estimated at $20.68$\%, and reduced to $19.78$\% at the start of experiment 2 (see Fig.~\ref{fig: InferenceResults}). As shown in Fig.~\ref{fig: InferenceExperimentsComparison}, the model updates trust distributions at each timestep. It reflects history-dependent continuity across all Stage-4 experiments. This aligns with the view that trust in human-machine systems should be continuously evolving~\cite{john1992trust}.

\subsubsection{Impact of Adverse Experiences}
In Stage-4, the parameters $\lambda^{*}$,~$\psi^{*}$, and $\theta^{*}$ were fixed. After the MLE process in Stage-3, the success and failure weights were $\omega^\mathtt{s}~=~3.7897$ and $\omega^\mathtt{f}~=~4.5390$, respectively. These results align with prior findings~\cite{yang2016users, guo2021modeling} that failures impact trust more than successes ($\omega^\mathtt{f}~\geq~\omega^\mathtt{s}$), as discussed in Section~\ref{sec: ExperientialObservations}).

\subsubsection{Nonlinearity in Trust Estimations}
In experiment 1, estimated trust decreased from $16.15$\% to below $10$\% by timestep~$10$, then increased to $20.68$\% by the termination timestep ($\mathtt{T}~=~20$) (see Fig.~\ref{fig: InferenceResults}). This pattern results from the piecewise beta reputation function, which depends on the threshold parameter $\varepsilon$ (Eq.~(\ref{eqn: beta_parameters})). A similar trend appears in experiment 2, where trust decreased until the timestep $4$ due to the robot approaching the obstacle too quickly, resulting in reward values below $\varepsilon$. In experiments 3 and 4, estimated trust remained low because the robot failed to complete the task. These results are consistent with prior work~\cite{guo2021modeling, azevedo2021real}.

\subsection{Comparison with Binary Performance-Based Trust Model}
To evaluate the granular trust estimation model, we conducted comparative experiments in Stage-4 using the binary performance-based trust model from~\cite{guo2021modeling}. As the source code was unavailable, we implemented the baseline and optimized both models via MLE using differential evolution~\cite{storn1997differential}. We calculated absolute error as the difference between the trust measurement and the trust estimation at the final timestep of each experiment (see Table~\ref{table: TableComparisonError}). A key advantage of our framework is its ability to provide trust estimations at fine-grained timescales without requiring labor-intensive, hand-crafted performance indicator design. However, a limitation is that the variance of the trust distribution tends to decrease over time, as the beta reputation model becomes more confident with each timestep. This suggests a future direction: exploring ways to regulate model confidence not only based on interaction number but also on contextual dynamics.
\afterpage{
    \begin{table}[t!]
        \centering
            \caption{Absolute Errors at the End of Each Inference Experiment.}
            \vspace{-3mm}
                \begin{tabular}{c|c|cc|cc}
                    \cmidrule{3-6}
                        \multicolumn{2}{c}{} & \multicolumn{4}{c}{\textit{Absolute Errors (\%)}} \\
                    \toprule
                        \multirow{3}{*}{\begin{tabular}[c]{@{}c@{}}Inference\\Number\end{tabular}} & \multirow{3}{*}{\begin{tabular}[c]{@{}c@{}}Self-Reported\\Human Trust\end{tabular}} & \multicolumn{2}{c|}{\begin{tabular}[c]{@{}c@{}}Binary\\Performance\\Based Model\end{tabular}} & \multicolumn{2}{c}{\begin{tabular}[c]{@{}c@{}}Granular\\Estimation of\\Trust (ours)\end{tabular}} \\
                    \cmidrule{3-6}
                        & & $\mu$ & $\sigma_{\mathtt{max}}^2$ & $\mu$ & $\sigma_{\mathtt{max}}^2$ \\
                    \midrule\midrule
                        $1$ & Moderate Distrust & $\mathbf{5.61}$ & $15.39$ & $9.31$ & $15.12$ \\
                        $2$ & Moderate Distrust & $\mathbf{0.09}$ & $11.12$ & $0.87$ & $6.74$ \\
                        $3$ & High Distrust & $19.17$ & $29.56$ & $\mathbf{8.91}$ & $12.78$ \\
                        $4$ & High Distrust & $18.61$ & $28.33$ & $\mathbf{3.52}$ & $6.35$ \\
                        $5$ & High Distrust & $18.17$ & $27.32$ & $\mathbf{0.18}$ & $2.35$ \\
                    \bottomrule
                \end{tabular}
            \vspace{1mm}
            \textit{}\\
            \textit{Bold values indicate which model has less error in each experiment.\\$\mu$ is the absolute error between self-reported trust and the mean of the estimated trust distribution; $\sigma_{\mathtt{max}}^2$ is the maximum variance of the error.}
        \label{table: TableComparisonError}
    \end{table}
}

\section{Conclusion} \label{sec: conclusion}
The proposed framework estimates trust at each timestep, offering a fine-grained view of trust dynamics. It also avoids manual performance metric design by using maximum entropy optimization to generate a continuous reward function.

While this work focuses on estimating trust at fine-grained timescales, it has three main limitations. First, the framework does not yet integrate trust estimation with robot control. However, this is a promising direction for future work. For instance, prior research~\cite{chen2020trust} adapted robot behavior based on trust, while other studies~\cite{de2020towards, li2024trust} explored trust repair strategies, such as apology, denial, and transparency. Since our framework estimates trust at each timestep, it offers the potential for real-time behavioral adaptation during a task. Second, the trust measurements in this study rely on self-reports, which limits validation to discrete snapshots rather than continuous measurement at every timestep. There are ongoing works in the development of real-time measures of behavioral, physiological, and neural signatures of trust~\cite{nam2020trust, wang2023human}, including electroencephalography, eye-tracking, and electrodermal activity measures~\cite{kohn2021measurement}. Third, using a Likert scale comes with a key limitation. A Likert scale is an ordinal measure, meaning the distance between response categories may vary across points and individuals~\cite{weijters2010stability}. This variability can introduce biases based on individual response styles. Despite this limitation, the Likert scale remains a pragmatic and cost-effective tool. It is widely used in psychology to capture subjectivity and has demonstrated strong predictive validity~\cite{eddy2019single}. An alternative to address this limitation is iterative optimization using preference-based data~\cite{campagna2024promoting}. In one study~\cite{bemporad2021global}, participants provided their preferences to improve the robot's performance and thereby increase human trust. However, this approach does not allow for the estimation of trust dynamics throughout the interaction. In contrast, accurately estimating the dynamics of human trust is necessary before applying strategies to mitigate undertrust and overtrust in the future~\cite{de2020towards, li2024trust, chen2020trust}. While this work differs from previous studies by focusing on the mathematical modeling of trust at fine-grained timescales, future research will explore ways to incorporate individual differences in experiments involving multiple participants.

\section*{Acknowledgement}
The authors would like to acknowledge Gibson Hu for technical support.

\bibliography{references}

\begin{thebibliography}{10}
\providecommand{\url}[1]{#1}
\csname url@rmstyle\endcsname
\providecommand{\newblock}{\relax}
\providecommand{\bibinfo}[2]{#2}
\providecommand\BIBentrySTDinterwordspacing{\spaceskip=0pt\relax}
\providecommand\BIBentryALTinterwordstretchfactor{4}
\providecommand\BIBentryALTinterwordspacing{\spaceskip=\fontdimen2\font plus
\BIBentryALTinterwordstretchfactor\fontdimen3\font minus \fontdimen4\font\relax}
\providecommand\BIBforeignlanguage[2]{{%
\expandafter\ifx\csname l@#1\endcsname\relax
\typeout{** WARNING: IEEEtran.bst: No hyphenation pattern has been}%
\typeout{** loaded for the language `#1'. Using the pattern for}%
\typeout{** the default language instead.}%
\else
\language=\csname l@#1\endcsname
\fi
#2}}

\bibitem{king2005getting}
B.~King-Casas, D.~Tomlin, C.~Anen, C.~F. Camerer, S.~R. Quartz, and P.~R. Montague, ``Getting to know you: reputation and trust in a two-person economic exchange,'' \emph{Science}, vol. 308, no. 5718, pp. 78--83, 2005.

\bibitem{ferrin2008takes}
D.~L. Ferrin, M.~C. Bligh, and J.~C. Kohles, ``It takes two to tango: An interdependence analysis of the spiraling of perceived trustworthiness and cooperation in interpersonal and intergroup relationships,'' \emph{Organizational behavior and human decision processes}, vol. 107, no.~2, pp. 161--178, 2008.

\bibitem{lewandowsky2000dynamics}
S.~Lewandowsky, M.~Mundy, and G.~Tan, ``The dynamics of trust: comparing humans to automation.'' \emph{Journal of Experimental Psychology: Applied}, vol.~6, no.~2, p. 104, 2000.

\bibitem{lewis2018role}
M.~Lewis, K.~Sycara, and P.~Walker, ``The role of trust in human-robot interaction,'' \emph{Foundations of trusted autonomy}, pp. 135--159, 2018.

\bibitem{gunia2024role}
A.~Gunia, ``The role of trust in human-machine interaction: Cognitive science perspective,'' in \emph{Artificial Intelligence, Management and Trust}.\hskip 1em plus 0.5em minus 0.4em\relax Routledge, 2024, pp. 85--126.

\bibitem{xu2015optimo}
A.~Xu and G.~Dudek, ``Optimo: Online probabilistic trust inference model for asymmetric human-robot collaborations,'' in \emph{Proceedings of the tenth annual ACM/IEEE international conference on human-robot interaction}, 2015, pp. 221--228.

\bibitem{nam2020trust}
C.~S. Nam and J.~B. Lyons, \emph{Trust in human-robot interaction}.\hskip 1em plus 0.5em minus 0.4em\relax Academic Press, 2020.

\bibitem{kaniarasu2013robot}
P.~Kaniarasu, A.~Steinfeld, M.~Desai, and H.~Yanco, ``Robot confidence and trust alignment,'' in \emph{2013 8th ACM/IEEE International Conference on Human-Robot Interaction (HRI)}.\hskip 1em plus 0.5em minus 0.4em\relax IEEE, 2013, pp. 155--156.

\bibitem{de2020towards}
E.~J. De~Visser, M.~M. Peeters, M.~F. Jung, S.~Kohn, T.~H. Shaw, R.~Pak, and M.~A. Neerincx, ``Towards a theory of longitudinal trust calibration in human--robot teams,'' \emph{International journal of social robotics}, vol.~12, no.~2, pp. 459--478, 2020.

\bibitem{yang2016users}
X.~J. Yang, C.~D. Wickens, and K.~Hölttä-Otto, ``How users adjust trust in automation: Contrast effect and hindsight bias,'' \emph{Proceedings of the Human Factors and Ergonomics Society Annual Meeting}, vol.~60, no.~1, pp. 196--200, 2016.

\bibitem{wang2023human}
Y.~Wang, F.~Li, H.~Zheng, L.~Jiang, M.~F. Mahani, and Z.~Liao, ``Human trust in robots: A survey on trust models and their controls/robotics applications,'' \emph{IEEE Open Journal of Control Systems}, 2023.

\bibitem{lewis2021deep}
M.~Lewis, H.~Li, and K.~Sycara, ``Deep learning, transparency, and trust in human robot teamwork,'' in \emph{Trust in human-robot interaction}.\hskip 1em plus 0.5em minus 0.4em\relax Elsevier, 2021, pp. 321--352.

\bibitem{wang2022computational}
Q.~Wang, D.~Liu, M.~G. Carmichael, S.~Aldini, and C.-T. Lin, ``Computational model of robot trust in human co-worker for physical human-robot collaboration,'' \emph{IEEE Robotics and Automation Letters}, vol.~7, no.~2, pp. 3146--3153, 2022.

\bibitem{williams2023computational}
K.~J. Williams, M.~S. Yuh, and N.~Jain, ``A computational model of coupled human trust and self-confidence dynamics,'' \emph{ACM transactions on human-robot interaction}, vol.~12, no.~3, pp. 1--29, 2023.

\bibitem{hancock2011meta}
P.~A. Hancock, D.~R. Billings, K.~E. Schaefer, J.~Y.~C. Chen, E.~J. de~Visser, and R.~Parasuraman, ``A meta-analysis of factors affecting trust in human-robot interaction,'' \emph{Human Factors}, vol.~53, no.~5, pp. 517--527, 2011.

\bibitem{schaefer2016measuring}
K.~E. Schaefer, ``Measuring trust in human robot interactions: Development of the “trust perception scale-hri”,'' in \emph{Robust intelligence and trust in autonomous systems}.\hskip 1em plus 0.5em minus 0.4em\relax Springer, 2016, pp. 191--218.

\bibitem{chen2018planning}
M.~Chen, S.~Nikolaidis, H.~Soh, D.~Hsu, and S.~Srinivasa, ``Planning with trust for human-robot collaboration,'' in \emph{Proceedings of the 2018 ACM/IEEE International Conference on Human-Robot Interaction}.\hskip 1em plus 0.5em minus 0.4em\relax Association for Computing Machinery, 2018, p. 307–315.

\bibitem{nam2019models}
C.~Nam, P.~Walker, H.~Li, M.~Lewis, and K.~Sycara, ``Models of trust in human control of swarms with varied levels of autonomy,'' \emph{IEEE Transactions on Human-Machine Systems}, vol.~50, no.~3, pp. 194--204, 2019.

\bibitem{guo2021modeling}
Y.~Guo and X.~J. Yang, ``Modeling and predicting trust dynamics in human-robot teaming: A bayesian inference approach,'' \emph{International Journal of Social Robotics}, vol.~13, no.~8, pp. 1899--1909, 2021.

\bibitem{bhat2022clustering}
S.~Bhat, J.~B. Lyons, C.~Shi, and X.~J. Yang, ``Clustering trust dynamics in a human-robot sequential decision-making task,'' \emph{IEEE Robotics and Automation Letters}, vol.~7, no.~4, pp. 8815--8822, 2022.

\bibitem{biyik2023active}
E.~Bıyık, N.~Huynh, M.~J. Kochenderfer, and D.~Sadigh, ``Active preference-based gaussian process regression for reward learning and optimization,'' \emph{The International Journal of Robotics Research}, 2023.

\bibitem{biyik2022learning}
E.~Bıyık, D.~P. Losey, M.~Palan, N.~C. Landolfi, G.~Shevchuk, and D.~Sadigh, ``Learning reward functions from diverse sources of human feedback: Optimally integrating demonstrations and preferences,'' \emph{The International Journal of Robotics Research}, vol.~41, no.~1, pp. 45--67, 2022.

\bibitem{finn2016guided}
C.~Finn, S.~Levine, and P.~Abbeel, ``Guided cost learning: Deep inverse optimal control via policy optimization,'' in \emph{Proceedings of The 33rd International Conference on Machine Learning}, ser. Proceedings of Machine Learning Research, vol.~48.\hskip 1em plus 0.5em minus 0.4em\relax PMLR, 2016, pp. 49--58.

\bibitem{li2024trust}
Y.~Li and F.~Zhang, ``Trust-preserved human-robot shared autonomy enabled by bayesian relational event modeling,'' \emph{IEEE Robotics and Automation Letters}, vol.~9, no.~11, pp. 10\,716--10\,723, 2024.

\bibitem{wu2017toward}
B.~Wu, B.~Hu, and H.~Lin, ``Toward efficient manufacturing systems: A trust based human robot collaboration,'' in \emph{2017 American Control Conference (ACC)}.\hskip 1em plus 0.5em minus 0.4em\relax IEEE, 2017, pp. 1536--1541.

\bibitem{john1992trust}
L.~John and N.~Moray, ``Trust, control strategies and allocation of function in human-machine systems,'' \emph{Ergonomics}, vol.~35, no.~10, pp. 1243--1270, 1992.

\bibitem{fleming2024metacognition}
S.~M. Fleming, ``Metacognition and confidence: A review and synthesis,'' \emph{Annual Review of Psychology}, vol.~75, no.~1, pp. 241--268, 2024.

\bibitem{yang2017evaluating}
X.~J. Yang, V.~V. Unhelkar, K.~Li, and J.~A. Shah, ``Evaluating effects of user experience and system transparency on trust in automation,'' in \emph{Proceedings of the 2017 ACM/IEEE International Conference on Human-Robot Interaction}.\hskip 1em plus 0.5em minus 0.4em\relax Association for Computing Machinery, 2017, p. 408–416.

\bibitem{josang2002beta}
A.~J{\o}sang and R.~Ismail, ``The beta reputation system,'' in \emph{Proceedings of the 15th bled electronic commerce conference}, 2002, pp. 324--337.

\bibitem{malle2020trust}
B.~F. Malle, K.~Fischer, J.~Young, A.~Moon, and E.~Collins, ``Trust and the discrepancy between expectations and actual capabilities,'' \emph{Human-robot interaction: Control, analysis, and design}, pp. 1--23, 2020.

\bibitem{lee2004trust}
J.~D. Lee and K.~A. See, ``Trust in automation: Designing for appropriate reliance,'' \emph{Human Factors}, vol.~46, no.~1, pp. 50--80, 2004.

\bibitem{ziebart2008maximum}
B.~D. Ziebart, A.~Maas, J.~A. Bagnell, and A.~K. Dey, ``Maximum entropy inverse reinforcement learning,'' in \emph{Proc. AAAI}, 2008, pp. 1433--1438.

\bibitem{campagna2024promoting}
G.~Campagna, M.~Lagomarsino, M.~Lorenzini, D.~Chrysostomou, M.~Rehm, and A.~Ajoudani, ``Promoting trust in industrial human-robot collaboration through preference-based optimization,'' \emph{IEEE Robotics and Automation Letters}, vol.~9, no.~11, pp. 9255--9262, 2024.

\bibitem{bemporad2021global}
A.~Bemporad and D.~Piga, ``Global optimization based on active preference learning with radial basis functions,'' \emph{Machine Learning}, vol. 110, no.~2, pp. 417--448, 2021.

\bibitem{swamy2023inverse}
G.~Swamy, D.~Wu, S.~Choudhury, D.~Bagnell, and S.~Wu, ``Inverse reinforcement learning without reinforcement learning,'' in \emph{International Conference on Machine Learning}.\hskip 1em plus 0.5em minus 0.4em\relax PMLR, 2023, pp. 33\,299--33\,318.

\bibitem{frey2024interpersonal}
V.~Frey and J.~Martinez, ``Interpersonal trust modelling through multi-agent reinforcement learning,'' \emph{Cognitive Systems Research}, vol.~83, p. 101157, 2024.

\bibitem{storn1997differential}
R.~Storn and K.~Price, ``Differential evolution--a simple and efficient heuristic for global optimization over continuous spaces,'' \emph{Journal of global optimization}, vol.~11, pp. 341--359, 1997.

\bibitem{azevedo2021real}
H.~Azevedo-Sa, S.~K. Jayaraman, C.~T. Esterwood, X.~J. Yang, L.~P. Robert~Jr, and D.~M. Tilbury, ``Real-time estimation of drivers’ trust in automated driving systems,'' \emph{International Journal of Social Robotics}, vol.~13, no.~8, pp. 1911--1927, 2021.

\bibitem{chen2020trust}
M.~Chen, S.~Nikolaidis, H.~Soh, D.~Hsu, and S.~Srinivasa, ``Trust-aware decision making for human-robot collaboration: Model learning and planning,'' \emph{J. Hum.-Robot Interact.}, vol.~9, no.~2, Jan. 2020.

\bibitem{kohn2021measurement}
S.~C. Kohn, E.~J. de~Visser, E.~Wiese, Y.-C. Lee, and T.~H. Shaw, ``Measurement of trust in automation: A narrative review and reference guide,'' \emph{Frontiers in Psychology}, vol. Volume 12 - 2021, 2021.

\bibitem{weijters2010stability}
B.~Weijters, M.~Geuens, and N.~Schillewaert, ``The stability of individual response styles.'' \emph{Psychological methods}, vol.~15, no.~1, pp. 96--110, 2010.

\bibitem{eddy2019single}
C.~L. Eddy, K.~C. Herman, and W.~M. Reinke, ``Single-item teacher stress and coping measures: Concurrent and predictive validity and sensitivity to change,'' \emph{Journal of School Psychology}, vol.~76, pp. 17--32, 2019.

\end{thebibliography}

\addtolength{\textheight}{-12cm}  

\end{document}